\documentclass[11pt]{article}

\pdfoutput=1

\usepackage[T1]{fontenc}
\usepackage[utf8]{inputenc}
\usepackage{times}
\usepackage{latexsym}
\usepackage{inconsolata}

\usepackage[final]{ACL2023} 

\usepackage{microtype}

\usepackage{amsmath}
\usepackage{amssymb}
\usepackage{amsthm}
\usepackage{amssymb}
\usepackage{graphicx} 

\usepackage{algorithm}
\usepackage{algpseudocode}

\usepackage{booktabs}
\usepackage{multirow}
\usepackage{graphicx} 
\usepackage{graphicx}
\usepackage{tikz}
\usepackage{tabularx} 

\usepackage{xcolor}

\usepackage{hyperref}

\usepackage{natbib}

\usepackage{xcolor}
\usepackage{array} 
\usepackage{tabularx} 
\usepackage{multirow}  
\usepackage{float}  
\usepackage{fancyvrb}
\usepackage{tabularx}  
\usepackage{threeparttable}
\usepackage{makecell}
\usepackage{booktabs}
\usepackage{adjustbox}

\usepackage{amsmath}

\title{Pat-DEVAL: Chain-of-Legal-Thought Evaluation for Patent Description}

\author{
  Yongmin Yoo \\
  Macquarie University \\
  \texttt{yooyongmin91@gmail.com} \\\And
   Kris W Pan \\
  Amazon \\
  \texttt{kriskwpan@gmail.com} \\\
}

\begin{document}
\maketitle
\begin{abstract} 
Patent descriptions must deliver comprehensive technical disclosure while meeting strict legal standards such as enablement and written description requirements. Although large language models have enabled end-to-end automated patent drafting, existing evaluation approaches fail to assess long-form structural coherence and statutory compliance specific to descriptions. We propose Pat-DEVAL, the first multi-dimensional evaluation framework dedicated to patent description bodies. Leveraging the LLM-as-a-judge paradigm, Pat-DEVAL introduces Chain-of-Legal-Thought (CoLT), a legally-constrained reasoning mechanism that enforces sequential patent-law-specific analysis. Experiments validated by patent expert on our Pap2Pat-EvalGold dataset demonstrate that Pat-DEVAL achieves a Pearson correlation of 0.69, significantly outperforming baseline metrics and existing LLM evaluators. Notably, the framework exhibits a superior correlation of 0.73 in Legal-Professional Compliance, proving that the explicit injection of statutory constraints is essential for capturing nuanced legal validity. By establishing a new standard for ensuring both technical soundness and legal compliance, Pat-DEVAL provides a robust methodological foundation for the practical deployment of automated patent drafting systems.
\end{abstract}
\section{Introduction}
As global patent applications reached a record high of approximately 3.7 million in 2024, marking the fifth consecutive year of growth~\citep{wipo2024indicators}, the demand for automated patent drafting tools to accelerate technological innovation and alleviate the burden on legal professionals has become more prominent than ever. In particular, the competition for innovation is intensifying, as evidenced by the United States Patent and Trademark Office receiving over 790,000 applications in 2024 alone~\citep{uspto2024dashboard}. Drafting patent descriptions is a highly knowledge-intensive process that requires not only a detailed technical exposition of an invention but also strict adherence to legal standards, such as enablement and sufficiency of disclosure~\citep{jiang2024survey}. Although recent studies have leveraged Large Language Models (LLMs) to transform scientific papers into patent descriptions, effective knowledge transfer remains challenging due to the fundamental rhetorical divergence between scientific papers, which emphasize experimental evidence, and patent descriptions, which prioritize legal protection and comprehensive technical disclosure~\citep{murray2005formal}.

\begin{figure}[t]
\centering
\includegraphics[width=0.99\linewidth,keepaspectratio]{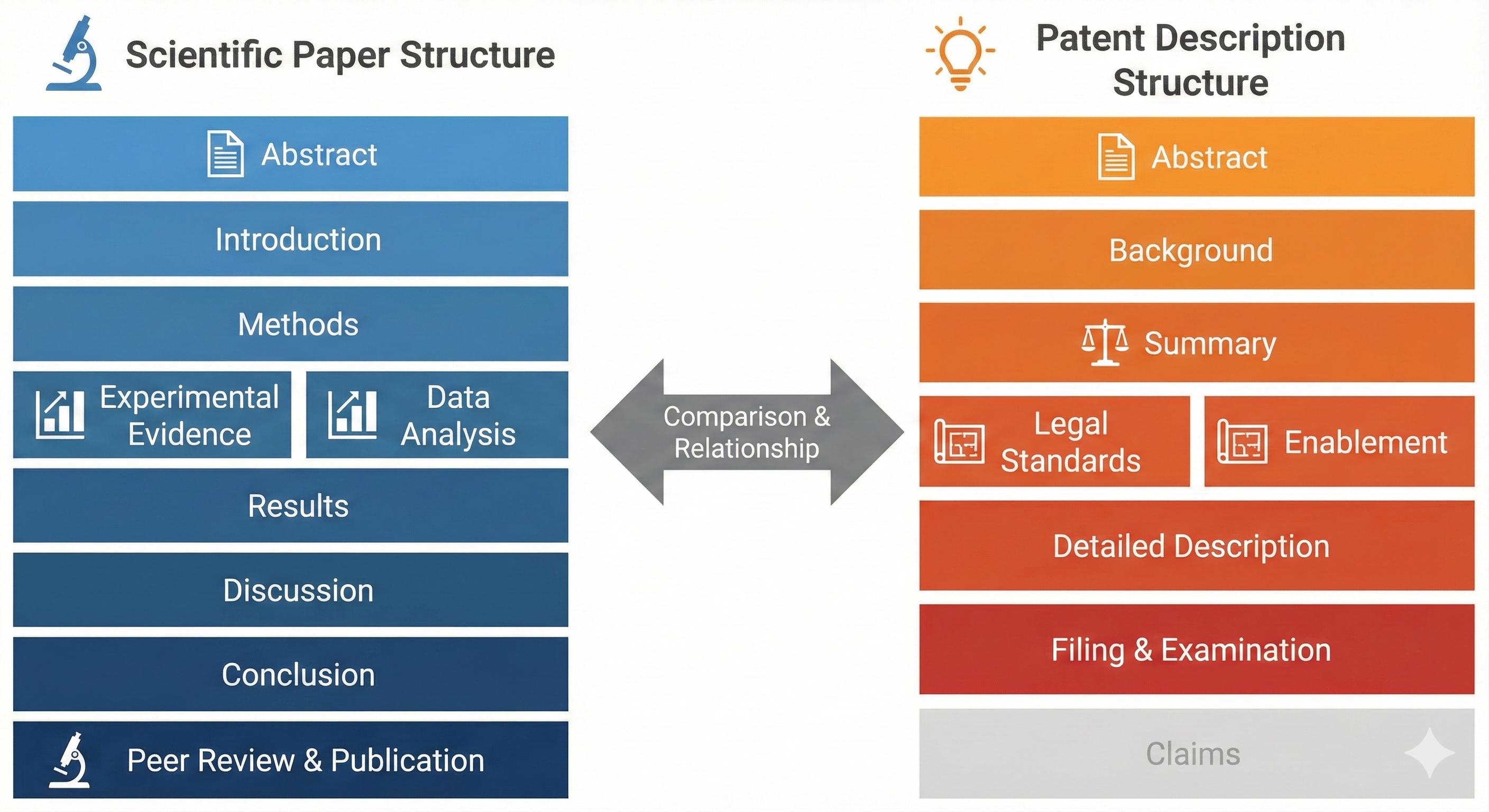}
\caption{Conceptual comparison of structural and rhetorical paradigms between scientific papers and patent descriptions.}
\label{fig:intro_overview}
\end{figure}

Despite these advancements in generative models, the evaluation of generated patent descriptions still largely relies on conventional NLP metrics, which fail to capture the structural coherence and rhetorical patterns essential to the legal domain~\citep{papineni2002bleu, lin2004rouge, zhang2019bertscore}. These traditional metrics focus on lexical overlap or embedding-based similarity, exhibiting fundamental limitations in assessing legal validity or cross-sectional logical flow in patent documents.

Recent LLM-as-a-judge approaches, such as PatentScore~\citep{yoo2025patentscore}, have advanced the evaluation of patent claims by addressing their unique logical consistency and scope definition. However, patent descriptions pose fundamentally different challenges: they form the bulk of the document and must provide sufficient technical disclosure to satisfy enablement and written description requirements under 35 U.S.C.~\S~112(a)~\citep{uscode112}. Evaluating descriptions thus requires assessing long-form structural completeness, cross-sectional logical flow, and reproducibility for a Person Having Ordinary Skill in the Art (PHOSITA), which are largely absent in claim evaluation. Consequently, existing frameworks cannot reliably determine whether a generated description is technically sound and legally compliant, hindering the practical deployment of automated patent drafting systems. This lack of specialized evaluation is becoming increasingly critical as the research community shifts toward end-to-end patent generation systems~\citep{knappich2025pap2pat, wang2024bautopatent}. These works have significantly advanced the field by proposing various automated drafting frameworks; however, they still lack a comprehensive methodological alternative for verifying the structural integrity and legal validity of the description bodies, leaving the generated outputs legally unverified.

To address these limitations, we propose Pat-DEVAL, a new structure-aware evaluation framework designed to bridge the gap between technical precision and legal requirements in patent descriptions. Pat-DEVAL utilizes LLMs as evaluators to analyze generated descriptions across four key dimensions: Technical Content Fidelity (TCF) and Data Precision (DP) to verify technical integrity, and Structural Coverage (SC) and Legal-Professional Compliance (LPC) to ensure adherence to specialized drafting conventions.

The contributions of this work are as follows:
\begin{itemize}
    \item \textbf{First Structure-Aware Framework for Patent Descriptions}: We introduce Pat-DEVAL, the first multi-dimensional framework dedicated to patent description, bridging the critical gap in evaluating long-form structural coherence and statutory enablement absent in prior claim-focused metrics.
    
    \item \textbf{Chain-of-Legal-Thought (CoLT)}: We propose a legally-constrained reasoning strategy that extends standard CoT by injecting statutory constraints and enforcing sequential patent-law analysis, reliably simulating the rigorous examination process of a PHOSITA.

    \item \textbf{Reference-Free Evaluation Paradigm}: We establish a source-anchored framework that evaluates descriptions directly against the source technology, accommodating the ``one-to-many'' nature of valid patent drafting where conventional reference-based metrics fail.

\end{itemize}
\section{Related Works}

\subsection{Automated Patent Generation}
Automated patent generation has evolved from initial statistical summarization models to full-scale specification drafting powered by LLMs. The release of BigPatent demonstrated that patent documents possess a significantly more complex discourse structure than standard news texts~\citep{sharma-etal-2019-bigpatent}. Consequently, to enhance abstractive summarization performance for long documents, pre-training objectives such as gap-sentences generation that mask and restore key sentences have been proposed~\citep{zhang2020pegasus}. 

Subsequent research has addressed the technical limitations associated with processing the extensive length of patent specifications. For instance, the introduction of BigBird~\citep{zaheer2020bigbird}, which utilizes a sparse attention mechanism to efficiently process contexts spanning thousands of tokens, has made it feasible to handle long-sequence patent texts. Building on these foundational technologies, prior attempts have fine-tuned GPT-2 to generate patent claims from invention summaries~\citep{lee-hsiang-2020-patent}; however, these efforts were largely confined to generating relatively short text segments.

More recently, end-to-end approaches attempting to generate entire patent descriptions from scientific papers have gained traction~\citep{knappich2025pap2pat}. While these studies have significantly improved the fluency of generated text by leveraging LLMs, they remain vulnerable in terms of structurally verifying the logical consistency and legal enablement required throughout the full-length specification body.

\subsection{LLM-based Evaluation Frameworks}

For decades, the evaluation of Natural Language Generation (NLG) has been dominated by n-gram overlap metrics such as BLEU, ROUGE, and METEOR~\citep{papineni2002bleu, lin2004rouge, banerjee2005meteor}. While computationally efficient, these metrics have been widely criticized for their inability to correlate well with human judgments regarding fluency and semantic equivalence. To address these shortcomings, embedding-based metrics like BERTScore~\citep{zhang2019bertscore} and MoverScore~\citep{zhao2019moverscore}, which utilize contextualized embeddings and Earth Mover's Distance, respectively, were introduced. Although these approaches mitigate the issues of lexical mismatch, they still fundamentally rely on surface-level or vector similarity, failing to capture complex logical inconsistencies or factual hallucinations inherent in long-form generation.

Recently, the paradigm has shifted toward LLM-as-a-Judge, leveraging the reasoning capabilities of LLMs to act as surrogate evaluators. For instance, GPTScore was proposed to evaluate text quality using the conditional probability of the model without fine-tuning~\citep{fu2023gptscore}, while G-Eval demonstrated that employing Chain-of-Thought (CoT) prompting can achieve higher alignment with human experts in summarization tasks~\citep{liu-etal-2023-geval}. Furthermore, specialized evaluator models such as Prometheus have been developed to enhance feedback reliability~\citep{kim2024prometheus}. However, these general-purpose frameworks are optimized for standard domains; they lack the domain-specific knowledge required to assess stringent legal criteria, such as the enablement requirement in patent law.

\subsection{Patent-specific Evaluation Metrics}

Prior to the generative AI era, patent quality assessment relied on bibliometric indicators and retrieval metrics. Studies often utilized statistical metadata, such as forward citations, family size, or claim counts, to infer the economic value or technological impact of a patent~\citep{squicciarini2013measuring}. In NLP, evaluation focused on the performance of downstream tasks, such as automated patent classification or information extraction, rather than the quality of the text itself~\citep{trappey2019review}. While valuable for analyzing existing databases, these methods offer no methodological basis for verifying the internal coherence or legal validity of newly generated patent drafts.

With the advent of automated drafting, the need for evaluating generated content has emerged. Most notably, PatentScore was proposed as a framework employing LLMs to evaluate the legal consistency and semantic fidelity of patent claims~\citep{yoo2025patentscore}. While marking a significant advancement for the evaluation of claims, which are concise and highly logical constructs defining the legal scope of protection, PatentScore does not address the fundamentally different challenges posed by patent descriptions.

In contrast to claims, which typically span only a few paragraphs and prioritize deductive logical rigor and the consistency of the antecedent basis, patent descriptions constitute the bulk of the document, often consisting of thousands of words. These descriptions serve the distinct legal purpose of providing sufficient technical disclosure to satisfy enablement and written description requirements under 35 U.S.C. § 112(a)~\citep{uscode112}. Therefore, evaluating descriptions demands assessment of long-form structural completeness, such as the coverage of mandatory sections including the background, summary, brief description of drawings, and detailed description. Furthermore, it requires analyzing cross-sectional logical flow and reproducibility for a Person Having Ordinary Skill in the Art (PHOSITA), aspects that are marginal or absent in claim evaluation. Consequently, existing frameworks leave a critical gap in evaluating the technical soundness and structural integrity of automatically generated patent specifications.
\section{Methodology: Pat-DEVAL}

\begin{figure*}[t]
    \centering    
    \includegraphics[width=0.99\textwidth,keepaspectratio]{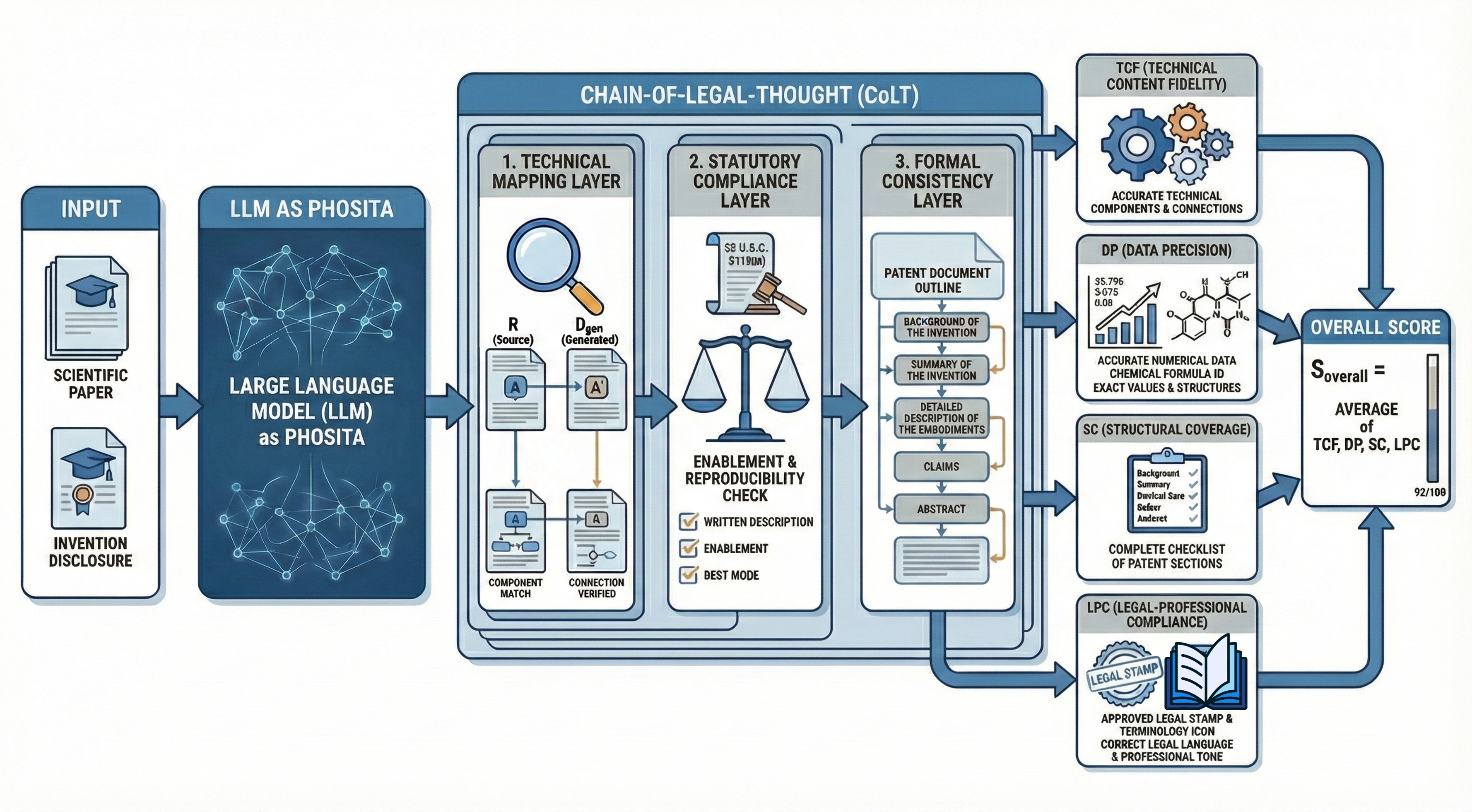}
    \caption{Overview of the Pat-DEVAL framework. The Chain-of-Legal-Thought (CoLT) mechanism enforces sequential reasoning through three patent-law-specific layers (Technical Mapping, Statutory Compliance, and Formal Consistency) using an LLM simulating a Person Having Ordinary Skill in the Art (PHOSITA). The process yields four dimension-specific scores (TCF, DP, SC, LPC), which are aggregated into an overall quality score.}
    \label{fig:workflow}
    \vspace{-4mm}
\end{figure*}

Existing Natural Language Generation (NLG) metrics and general-purpose LLM-based evaluation frameworks exhibit fundamental limitations when assessing legally binding documents. A patent specification is not merely a sequence of information but a sophisticated logical construct that transforms technical ideas into legal rights. To bridge this gap, we propose Pat-DEVAL, a framework that models the professional judgment process of patent examiners into a computable format. The core of our methodology lies in the Chain-of-Legal-Thought (CoLT) mechanism, which enforces the alignment of legal domain knowledge with the inference trajectory of the neural network.

\subsection{Framework Overview}
Pat-DEVAL is designed to evaluate not only whether the generated description $D_{gen}$ accurately reflects the source technology $R$, defined here as the full academic paper encompassing the abstract, methodology, and experiments, but also whether it provides sufficient technical disclosure to support the scope of rights. To achieve this, we simulate the LLM evaluator $\mathcal{M}$ as a Person Having Ordinary Skill in the Art (PHOSITA). The evaluation constitutes a structured inference pipeline where $\mathcal{M}$ conducts a legal reasoning process via CoLT and subsequently derives quantitative scores across four distinct dimensions.

\subsection{Chain-of-Legal-Thought (CoLT)}

\begin{table*}[t]
\small
\centering
\begin{tabular}{p{0.16\textwidth} p{0.22\textwidth} p{0.22\textwidth} p{0.28\textwidth}}
\hline
\textbf{Layer} & \textbf{Source Tech ($R$)} & \textbf{Gen. Patent ($D_{gen}$)} & \textbf{CoLT Reasoning (Internal)} \\ \hline
\textbf{Technical Mapping} & Proposed a "Dynamic Gating Module" using Gumbel-Softmax for sparsity. & Mentions a "filtering unit" to reduce computation. & \textit{Omission identified}: The specific mathematical mechanism (Gumbel-Softmax) is missing; replaced by a generic term. \\ \hline
\textbf{Statutory Compliance} & Hyperparameter $\lambda$ set to 0.01 for stable convergence. & Suggests using "an appropriate value" for stability. & \textit{Enablement Issue}: "Appropriate value" is too vague for a PHOSITA to reproduce results. \\ \hline
\textbf{Formal Consistency} & Includes a flowchart in Figure 3 illustrating the process. & Brief Description mentions Fig. 3, but the Detailed Description omits it. & \textit{Cross-sectional Inconsistency}: Figure 3 is defined in the list of drawings but never described in the body text. \\ \hline
\end{tabular}
\caption{A representative example of CoLT layers identifying specific deficiencies during the evaluation process.}
\label{tab:colt_example}
\end{table*}

Chain-of-Legal-Thought (CoLT) represents a fundamental departure from standard Chain-of-Thought (CoT) prompting. Unlike multi-turn conversational approaches that incur high latency, CoLT employs a single-pass, constraint-enforced prompting strategy. It introduces a structured inference trajectory that mandates sequential progression through patent-law-specific analytical layers within a single generation pass.

We explicitly inject external statutory constraints, such as 35 U.S.C. \S 112(a)~\citep{uscode112}, into the system prompt, preventing the model from outputting a final score until the reasoning trace is fully articulated. To operationalize this, CoLT requires the evaluator LLM to sequentially navigate three domain-specific analytical layers before scoring. Table~\ref{tab:colt_example} illustrates how these layers detect specific technical and legal discrepancies between the source technology $R$ and the generated output $D_{gen}$.

\paragraph{Technical Mapping Layer} This layer conducts a granular, element-by-element comparative analysis between the source document $R$ and the generated description $D_{gen}$. It identifies factual hallucinations or omissions that serve as the evidential foundation for fidelity scoring.

\paragraph{Statutory Compliance Layer} By simulating the perspective of a PHOSITA, this layer evaluates the enablement requirement. It checks whether the invention can be reproduced solely from $D_{gen}$ without undue experimentation, penalizing missing embodiments or implementation details.

\paragraph{Formal Consistency Layer} This layer assesses cross-sectional logical coherence and adherence to professional rhetorical conventions, ensuring the document meets the structural standards required for patentability.

%%%%%%%%%%%%%%%%%%%%%%%%%%%%%%%%%%%%%%%%%%%%%%%%%%%%%%%%%%%%%%%%%%%%%%%
%%%%%%%%%%%%%%%%%%%%%%%%%%%%%%%%%%%%%%%%%%%%%%%%%%%%%%%%%%%%%%%%%%%%%%%
%%%%%%%%%%%%%%%%%%%%%%%%%%%%%%%%%%%%%%%%%%%%%%%%%%%%%%%%%%%%%%%%%%%%%%%

\subsection{Evaluation Taxonomy}
The reasoning results derived from CoLT are crystallized into four formalized metrics. Each dimension is evaluated on a 1-to-5 Likert scale, where 1 represents a fatal failure to meet legal or technical standards, and 5 represents professional-grade drafting quality.

\begin{itemize}
    \item \textbf{Technical Content Fidelity (TCF):} Measures the distortion or omission of the source technology's core mechanisms.
    \item \textbf{Data Precision (DP):} Evaluates the exact match of experimental values, reference numerals, and chemical formulas.
    \item \textbf{Structural Coverage (SC):} Determines the completeness of mandatory sections required by patent statutes.
    \item \textbf{Legal-Professional Compliance (LPC):} Verifies adherence to the enablement requirement and appropriate legal phraseology.
\end{itemize}

Formally, each score $s_i \in \{1, \dots, 5\}$ and its corresponding rationale $\rho_i$ are derived via the function $f_{CoLT}$, taking the prompt $P$, legal constraints $L$, and the source and generated documents as inputs:
\begin{equation}
s_i, \rho_i = f_{CoLT}(P, L, R, D_{gen})
\end{equation}
The overall Pat-DEVAL score is computed as the unweighted average of the four dimension scores. While legal validity is often binary, we adopt this averaging approach to provide a holistic assessment of generation quality across different aspects of patent drafting. The detailed system prompts, including the specific statutory constraints and the 5-point grading rubrics for each dimension, are provided in Appendix \ref{sec:appendix_prompts}. 
\section{Experiments}
In this section, we describe the experimental setup and evaluation results designed to validate the efficacy of the Pat-DEVAL framework. The primary objective is to assess the accuracy of the framework in measuring the legal and technical quality of generated patent specifications and to quantify its correlation with human expert judgment. To this end, we construct a benchmark dataset encompassing diverse technical domains and conduct a comparative analysis against both traditional n-gram metrics and state-of-the-art LLM-based evaluators. We specifically aim to highlight the limitations of existing approaches that rely solely on textual similarity and to demonstrate that the CoLT mechanism is an essential component for capturing the unique structural and statutory constraints of patent descriptions. Furthermore, qualitative assessment data from certified patent professionals serve as the gold standard to comprehensively examine the reliability and explainability of Pat-DEVAL.

\subsection{Experimental Setup \& Dataset}

To evaluate the efficacy of Pat-DEVAL, we established a robust experimental pipeline comprising data curation, candidate generation, and expert annotation.

\paragraph{Dataset Construction}
We constructed Pap2Pat-EvalGold, a high-precision dataset derived from the Pap2Pat~\citep{knappich2025pap2pat} corpus. To ensure ground-truth reliability, we filtered the corpus using Sentence-BERT to measure the semantic similarity between titles and abstracts by setting a minimum BERTScore threshold of 0.8. We further verified the author overlap ratio to ensure it meets or exceeds 0.5, resulting in the selection of 146 high-quality paper-patent pairs.

\paragraph{Human Annotation}
To establish a high-fidelity gold standard, we conducted a comprehensive manual assessment of the entire Pap2Pat-EvalGold dataset ($N=146$), where each sample was independently scored by three certified patent professionals. We validated the reliability of these judgments using Krippendorff's Alpha and the Intraclass Correlation Coefficient (ICC). The analysis yielded an average Alpha of 0.764 and an ICC of 0.81. This ICC score exceeds the 0.80 threshold typically required for excellent reliability. Notably, achieving such high consensus is significant given the inherent linguistic ambiguity and complex legal-technical requirements of patent descriptions. This robust agreement confirms that our Pap2Pat-EvalGold dataset serves as a high-fidelity benchmark, minimizing the risk of individual bias in the subsequent correlation analysis with automated metrics.

\paragraph{Candidate Generation}
Using the academic papers from this dataset as source inputs, we employed Llama-3.1-70B to generate the corresponding patent descriptions. To ensure reproducibility and minimize stochasticity, all generation tasks were conducted with the decoding temperature fixed at 0.3 and the Top-k parameter set to 10. These generated descriptions serve as the target candidates for evaluation.

\subsection{Main Results}
\label{sec:main_results}

Table~\ref{tab:main_results} presents the Pearson correlation coefficients between automated evaluation metrics and human expert judgments across the four dimensions of patent quality. The results demonstrate that Pat-DEVAL achieves the highest alignment with professional standards by reaching a correlation of 0.69. This performance significantly outperforms all baseline methods with statistical significance at a p-value less than 0.01.

\begin{table}[ht]
\centering
\small
\renewcommand{\arraystretch}{1.2}
\resizebox{\columnwidth}{!}{
\begin{tabular}{lccccc}
\hline
\textbf{Metrics} & \textbf{TCF} & \textbf{DP} & \textbf{SC} & \textbf{LPC} & \textbf{Avg.} \\ \hline
\multicolumn{6}{l}{\textit{\textbf{Traditional \& Embedding Metrics}}} \\
\quad BLEU & 0.12 & 0.08 & 0.05 & 0.03 & 0.07 \\
\quad ROUGE-L & 0.15 & 0.10 & 0.07 & 0.04 & 0.09 \\
\quad BERTScore & 0.31 & 0.27 & 0.18 & 0.15 & 0.23 \\ \hline
\multicolumn{6}{l}{\textit{\textbf{LLM-as-a-Judge (Backbone: Qwen3-32B)}}} \\
\quad GPTScore & 0.38 & 0.35 & 0.29 & 0.22 & 0.31 \\
\quad Standard CoT & 0.45 & 0.41 & 0.38 & 0.32 & 0.39 \\
\quad Prometheus-2 & 0.51 & 0.48 & 0.44 & 0.39 & 0.46 \\
\quad G-Eval & 0.58 & 0.54 & 0.51 & 0.45 & 0.52 \\ \hline
\multicolumn{6}{l}{\textit{\textbf{Proposed Framework (Backbone: Qwen3-32B)}}} \\
\quad \textbf{Pat-DEVAL (Ours)} & \textbf{0.68} & \textbf{0.72} & \textbf{0.64} & \textbf{0.73} & \textbf{0.69} \\ \hline
\end{tabular}%
}
\caption{Pearson correlations ($r$) with human judgments on \texttt{Pap2Pat-EvalGold} ($N=146$). Using a controlled Qwen3-32B backbone, Pat-DEVAL significantly outperforms baselines, particularly in the Legal-Professional Compliance (LPC) dimension ($p < 0.01$).}
\label{tab:main_results}
\end{table}

\paragraph{Failure of Surface-Level Metrics}
Traditional n-gram metrics such as BLEU and ROUGE-L exhibit negligible correlation with human experts, yielding coefficients of 0.09 or lower. This confirms that lexical overlap is an ineffective proxy for patent quality, as valid patent descriptions for the same invention often employ diverse terminology and structures, reflecting a \textit{one-to-many} mapping inherent in patent drafting. While embedding-based metrics like BERTScore show a marginal improvement by capturing semantic similarity and reaching a correlation of 0.23, they remain insufficient for assessing the structural coherence and statutory compliance required for patentability.

\paragraph{Limitations of Generic and Fine-Tuned LLMs}
While LLM-based evaluators generally outperform traditional metrics, generic prompting strategies reveal distinct limitations within the legal domain. Methods such as Standard CoT and GPTScore struggle to distinguish between linguistically fluent text and legally enabling disclosure, recording correlations of 0.39 and 0.31, respectively. A significant observation concerns Prometheus-2, a model explicitly fine-tuned for evaluation tasks. Despite its specialized training, Prometheus-2 recorded a correlation of 0.46, which is lower than the 0.52 achieved by the prompting-based G-Eval framework. This discrepancy suggests that general-purpose evaluation tuning does not automatically transfer to the patent domain without specific legal grounding. Consequently, these models exhibit limited performance in the Legal-Professional Compliance dimension as evidenced by correlation scores ranging from 0.39 to 0.45.

\paragraph{Superiority of Pat-DEVAL}
Pat-DEVAL demonstrates superior alignment with human judgments, surpassing the strongest baseline, G-Eval, by a margin of 0.17. The performance gap is most pronounced in the Legal-Professional Compliance dimension, where Pat-DEVAL achieves a correlation of 0.73 compared to the 0.45 recorded by G-Eval. This validates the hypothesis that the Chain-of-Legal-Thought mechanism successfully simulates the reasoning process of a PHOSITA, enabling the model to rigorously evaluate enablement and written description requirements. Since all LLM-based evaluators utilized the identical Qwen3-32B backbone, this performance gain is attributable to the proposed domain-specific methodology rather than differences in underlying model capacity.

\subsection{Ablation Study}

To verify the individual contributions of the key components constituting the Pat-DEVAL framework, we conducted an ablation study using the Qwen3-32B backbone model. Table~\ref{tab:ablation} summarizes the performance changes observed upon removing the CoLT logic, the PHOSITA persona, and the dimensional decomposition structure.

\begin{table}[ht]
\centering
\small
\renewcommand{\arraystretch}{1.2}
\resizebox{\columnwidth}{!}{
\begin{tabular}{lccccc}
\hline
\textbf{Configuration} & \textbf{TCF} & \textbf{DP} & \textbf{SC} & \textbf{LPC} & \textbf{Avg.} \\ \hline
\textbf{Pat-DEVAL (Full)} & \textbf{0.68} & \textbf{0.72} & \textbf{0.64} & \textbf{0.73} & \textbf{0.69} \\ \hline
\quad w/o CoLT Logic & 0.49 & 0.45 & 0.41 & 0.35 & 0.43 \\
\quad w/o PHOSITA Persona & 0.61 & 0.65 & 0.58 & 0.62 & 0.62 \\
\quad w/o Decomposition & 0.52 & 0.49 & 0.46 & 0.41 & 0.47 \\ \hline
\end{tabular}%
}
\caption{Ablation analysis on \texttt{Pap2Pat-EvalGold} (Backbone: Qwen3-32B). Values denote Pearson correlations ($r$) with human judgments, highlighting the contribution of each framework component.}
\label{tab:ablation}
\end{table}

\paragraph{Impact of Chain-of-Legal-Thought}
The most drastic performance degradation was observed when the legal reasoning mechanism was removed. In the configuration lacking CoLT logic, we replaced the specific statutory reasoning steps based on 35 U.S.C. with the generic prompt say Let's think step by step. Consequently, the average correlation plummeted from 0.69 to 0.43. Notably, the correlation in the Legal-Professional Compliance dimension dropped to 0.35, a level comparable to the Standard CoT baseline. This strongly suggests that the performance gain of Pat-DEVAL stems not from the general reasoning capability of the LLM but from the explicit injection of domain-specific legal knowledge.

\paragraph{Role of PHOSITA Persona}
When the evaluator was not assigned the role of acting as a Person Having Ordinary Skill in the Art, the average correlation decreased moderately from 0.69 to 0.62. While the model retained some ability to assess technical content, the decline is evident. Explicitly defining the expert perspective contributes to aligning the judgments of the model with the standard of review used by patent attorneys, particularly when judging the appropriateness of technical terminology and the level of invention.

\paragraph{Effect of Dimensional Decomposition}
Finally, we validated the efficacy of stratifying patent quality into four dimensions. When instructed to assign a holistic score without dimensional distinction, the correlation reached only 0.47. This implies that the quality of a patent specification is too complex to be compressed into a single scalar value. The strategy of decomposing the evaluation target into specific dimensions ensures that the model focuses on each aspect, thereby serving as an essential mechanism to reduce hallucinations common in abstract scoring and to improve precision.

\section{Key Findings and Discussion}

Based on the experimental results and ablation studies, we derive the following key academic insights regarding the evaluation of automated patent generation.

\paragraph{Decoupling of Lexical Similarity and Legal Validity}
Our results show that traditional n-gram metrics (BLEU, ROUGE) fail to achieve any meaningful correlation with human expert judgment. This confirms that lexical overlap is a poor proxy for patent quality. Unlike general-purpose summarization, patent descriptions exhibit a one-to-many mapping, where the same inventive concept can be articulated through diverse linguistic structures and strategic legal terminologies. Consequently, high textual similarity does not guarantee technical accuracy or statutory compliance, necessitating a domain-specific evaluation framework like Pat-DEVAL.

\paragraph{Efficacy of Explicit Statutory Constraints}
The most significant finding from the ablation study is the critical role of the Chain-of-Legal-Thought (CoLT) mechanism. The sharp decline in correlation within the Legal-Professional Compliance (LPC) dimension (from 0.73 to 0.35) when CoLT was removed demonstrates that the general reasoning capabilities of LLMs are insufficient for high-stakes legal assessments. Explicitly injecting statutory requirements, such as 35 U.S.C. §112, is essential for transforming a general-purpose evaluator into a specialized legal judge capable of assessing enablement and written description standards.

\paragraph{Expert Alignment via PHOSITA Persona}
Assigning the PHOSITA persona to the evaluator significantly improved alignment with patent professionals. This suggests that the model's judgment is sensitive to the defined expert perspective, allowing it to move beyond linguistic fluency to evaluate the appropriateness of technical disclosures and the level of ordinary skill in the art. This finding implies that providing the correct professional context is as crucial as the underlying model's capacity in specialized domain evaluation.

\paragraph{Precision through Dimensional Decomposition}
The transition from holistic scoring to a multi-dimensional approach (TCF, DP, SC, and LPC) led to a substantial performance gain ($r=0.47 \rightarrow 0.69$). This decomposition acts as a grounding mechanism that reduces the risk of abstract hallucinations. By forcing the model to generate a structured reasoning trace for each specific legal and technical dimension, Pat-DEVAL ensures both higher precision and better explainability of the final assessment.
\section{Conclusion}

In this paper, we introduced Pat-DEVAL, the first multi-dimensional evaluation framework specifically engineered for the body of patent descriptions. By simulating a Person Having Ordinary Skill in the Art (PHOSITA) through the Chain-of-Legal-Thought (CoLT) mechanism, we successfully bridged the critical gap between technical summarization and legal validity assessment. Unlike traditional surface-level metrics that rely on lexical overlap, Pat-DEVAL enforces a structured reasoning trajectory that incorporates explicit statutory constraints, such as the enablement and written description requirements under 35 U.S.C. 112(a).

Our experimental results on the Pap2Pat-EvalGold dataset demonstrate that Pat-DEVAL achieves a Pearson correlation of 0.69 with human expert judgments, significantly outperforming all baseline metrics. The ablation study further underscores the indispensability of the CoLT logic, particularly in the Legal-Professional Compliance (LPC) dimension, where it recorded a correlation of 0.7. These findings confirm that ensuring the legal soundness of automated patent drafting requires more than general linguistic fluency; it necessitates the rigorous injection of domain-specific legal reasoning.

As LLM-driven patent generation shifts toward end-to-end production, Pat-DEVAL establishes a new standard for verifying the structural integrity and technical fidelity of generated specifications. This framework provides a robust methodological foundation to ensure that automated patent drafts are both technically accurate and legally valid.
\section*{Limitation}

This study focuses exclusively on the description body of patent specifications to evaluate technical fidelity and statutory enablement, deliberately excluding claims from the scope of evaluation. This separation is grounded in the fundamental divergence between the linguistic and legal attributes required for claims versus those for descriptions. While claims necessitate an analysis of logical inclusion and the clarity of legal boundaries, the description body must encompass extensive technical disclosures and provide sufficient information for a Person Having Ordinary Skill in the Art (PHOSITA) to reproduce the invention. Thus, our framework serves as a critical and non-redundant complement to existing claim-centric evaluation methodologies by establishing the first dedicated validation mechanism for the structural and legal integrity of patent descriptions.

Furthermore, the Pap2Pat-EvalGold dataset, developed to validate these sophisticated evaluation criteria, consists of 146 sample pairs, which is a scale relatively modest compared to large-scale general-purpose LLM datasets. However, this was a strategic prioritization of qualitative fidelity over quantitative volume, dictated by the specialized nature of the patent domain. Assessing complex legal requirements such as enablement cannot be outsourced to general crowdsourcing because it necessitates the rigorous judgment of certified patent professionals with domain-specific expertise. Despite the limited sample size, the high Intraclass Correlation Coefficient (ICC) of 0.81 underscores the robustness of our dataset, positioning it as a reliable gold standard for benchmarking automated patent evaluation models. Future research will aim to integrate Pat-DEVAL with claim-evaluation models and expand domain coverage through a human-in-the-loop pipeline to develop a holistic and end-to-end patent evaluation system.

\bibliography{anthology,custom}
\bibliographystyle{acl_natbib}

\appendix
\appendix
\section{Pat-DEVAL Prompt Details}
\label{sec:appendix_prompts}

\subsection{System Prompt Design for CoLT}
The following system prompt is utilized to simulate the PHOSITA evaluator. To ensure the evaluation is grounded in legal standards, we enforce a structured reasoning process that must be completed prior to the assignment of quantitative scores.

\begin{table*}[h]
\small
\centering
\begin{tabular}{p{0.95\textwidth}}
\hline
\textbf{Full System Prompt for Pat-DEVAL Evaluator} \\ \hline
\textbf{[Role Definition]} \\
You are a Senior Patent Examiner with expertise in the relevant technical domain. Evaluate the generated patent description ($D_{gen}$) based on the source technology ($R$) provided in the academic paper. You must act as a \textbf{Person Having Ordinary Skill in the Art (PHOSITA)} as defined in patent jurisprudence. \\
\\
\textbf{[Requirement: Chain-of-Legal-Thought (CoLT)]} \\
Before providing scores, you must explicitly document your reasoning across the following three layers: \\
\\
\textbf{1. Technical Mapping Layer}: Conduct an element-by-element comparison between $R$ and $D_{gen}$. Identify all core novel mechanisms, experimental data, and technical embodiments. Note any factual hallucinations, distortions, or omissions. \\
\\
\textbf{2. Statutory Compliance Layer}: Evaluate whether $D_{gen}$ satisfies the \textbf{Enablement Requirement (35 U.S.C. \S 112(a))}. Determine if the disclosure is sufficient for a PHOSITA to reproduce the invention without "undue experimentation." Also, check for the \textbf{Written Description Requirement} to ensure the inventor had possession of the invention. \\
\\
\textbf{3. Formal Consistency Layer}: Assess the structural integrity of the document (e.g., Background, Summary, Detailed Description) and its adherence to professional patent drafting conventions and legal terminology. \\
\\
\textbf{[Scoring and Output Format]} \\
After the analysis, provide scores (1-5) for: (1) Technical Content Fidelity (TCF), (2) Data Precision (DP), (3) Structural Coverage (SC), and (4) Legal-Professional Compliance (LPC). \\
\\
\textbf{Output Template:} \\
- \textbf{Reasoning Trace}: [Insert Step 1, 2, 3 analysis here] \\
- \textbf{Scores}: TCF: [X], DP: [X], SC: [X], LPC: [X] \\
- \textbf{Final Rationale}: [Brief justification for the final scores] \\ \hline
\end{tabular}
\caption{The complete system prompt for the Pat-DEVAL framework. The prompt is designed to minimize heuristic biases by mandating a sequential legal reasoning trace (CoLT) before quantitative assessment.}
\label{tab:full_prompt}
\end{table*}

\subsection{Evaluation Rubrics}
To maintain inter-evaluator consistency, the LLM is instructed to follow the 5-point Likert scale rubrics defined below. These criteria apply across all four dimensions (TCF, DP, SC, LPC):

\paragraph{Score 5 (Professional)} The description aligns perfectly with the source technology $R$. It demonstrates flawless legal structure, professional terminology, precise data reproduction, and complete enablement of all \textbf{disclosed embodiments}.

\paragraph{Score 4 (High Quality)} Most core mechanisms and experimental values are accurately captured with minor, non-critical omissions. It satisfies the enablement requirement and maintains a high degree of formal consistency across all mandatory sections.

\paragraph{Score 3 (Acceptable)} The primary invention is disclosed, but the description may lack specific implementation details, contain minor numerical discrepancies, or exhibit terminological inconsistencies that would require slight revision before filing.

\paragraph{Score 2 (Mediocre)} Significant technical gaps or data ambiguities exist. A PHOSITA would likely require undue experimentation to reproduce the invention based solely on the provided text, or key sections (e.g., Detailed Description) are underdeveloped.

\paragraph{Score 1 (Fatal)} The document contains critical factual hallucinations or fails to describe the core invention. It does not meet the statutory requirements under 35 U.S.C. \S 112 due to severe structural or technical deficiencies.

\end{document}